\documentclass[letterpaper, 10 pt, conference]{ieeeconf}  

\usepackage{soul, xcolor}
\usepackage{booktabs, comment, makecell}
\usepackage{colortbl}
\usepackage{graphicx}
\usepackage[hidelinks]{hyperref}

\newcommand{\lo}{\textsc{low}}
\newcommand{\mi}{\textsc{mid}}
\newcommand{\hi}{\textsc{high}}
\definecolor{myblue}{HTML}{cfedff}
\definecolor{mygrey}{HTML}{dee0df}
\definecolor{mygrey2}{HTML}{f0f2f1}

\IEEEoverridecommandlockouts                              

\title{\LARGE \bf
The Influence of Human-like Appearance on Expected Robot Explanations
}

\author{Hana Kopecka$^{1}$ and Jose Such$^{2}$
\thanks{*This work was supported by UK Research and Innovation [grant number EP/S023356/1], in the UKRI Centre
for Doctoral Training in Safe and Trusted Artificial Intelligence (www.safeandtrustedai.org), the INCIBE's strategic SPRINT (Seguridad y Privacidad en Sistemas con Inteligencia Artificial) C063/23 project with funds from the EU-NextGenerationEU through the Spanish government's Plan de Recuperación, Transformación y Resiliencia, the GENERALITAT VALENCIANA project PROMETEO CIPROM/2023/23, and the Digital Humanism Fellowship at the Institute for Human Sciences (IWM), funded by the Austrian Federal Ministry for Climate Action, Environment, Energy, Mobility, Innovation and Technology (BMK)}
\thanks{$^{1}$Hana Kopecka is with VRAIN, Universitat Politècnica de València, Camino de Vera, s/n 46022 Valencia, Spain
        {\tt\small hkopeck1@upv.edu.es}}%
\thanks{$^{2}$Jose Such is with INGENIO (CSIC-Universitat Politecnica de Valencia),
        Cami de Vera, s/n 46022 Valencia, Spain
        {\tt\small jose.such@csic.es}}%
}

\begin{document}

\maketitle
\thispagestyle{empty}
\pagestyle{empty}

\begin{abstract}

A robot's appearance is a known factor influencing user's mental model and human-robot interaction, that has not been studied in the context of its influence in expected robot explanations. In this study, we investigate whether and to what extent the human-like appearance of robots elicits anthropomorphism, which is conceptualised as an attribution of mental capacities, and how the level of anthropomorphism is revealed in explanations that people expect to receive. We designed a between-subject study comprising conditions with visual stimuli of three domestic service robots with varying human-like appearance, and we prompted respondents to provide explanations they would expect to receive from the robot for the same robot actions. 
We found that most explanations were anthropomorphic across all conditions. However, there is a positive correlation between the anthropomorphic explanations and human-like appearance. We also report on more nuanced trends observed in non-anthropomorphic explanations and trends in robot descriptions.

\end{abstract}

\section{Introduction}

Explainability is critical in human-AI and human-robot interactions \cite{Setchi2020,sado2023},
to facilitate  transparency, fairness, and accountability~\cite{wachter}, but also to help in building an appropriate mental model of the robot and its skills to
support human-robot collaboration and prevent failure~\cite{Anjomshoae1303810}, and to help address situations in which the robot fails to achieve its goal or behaves unexpectedly \cite{sado2023}.
However, there are several challenges in achieving explainable robotics such as deciding what needs to be explained to a particular individual in a given context as well as the technical challenge in developing algorithms that can generate an appropriate explanation \cite{Setchi2020}, all of which requires an interdisciplinary effort \cite{Papagni2021}. One of the challenges of explainable robotics is people forming incorrect mental models about the robot's mental (perceptual and cognitive) capacities which hinders users' ability to predict the robot's behaviour and it also results in communication failures \cite{Thellman2021}. A known factor that influences the attribution of mental capacities to robots is human-like appearance \cite{looser2010,Martini2016}.
While literature documents that an increasing human-like appearance elicits higher attribution of mental capacities to robots, e.g. \cite{looser2010,Martini2016}, many studies rely on participants' explicit rating of the degree to which they perceive the robot to be endowed with mental capacities \cite{Banks2021}. However, there is also evidence that having a conscious belief that the robot (or computer) has a mind is not the same as unconsciously perceiving and treating the robot as having a mind \cite{Banks2020,Kim2012}. To our knowledge, this implicit perception of a robot's mental capacities has not been studied in the context of explainable robotics, and in particular, whether people expect robots to reference its mental capacities in explanations of their actions and if this is influenced by the robots' appearance. 

In this explorative study, we investigate the level of anthropomorphism (i.e. the attribution of human mental capacities \cite{Kuhne2023}) revealed in expected explanations by three domestic service robots of different levels of human-like appearance. To this end, we designed a study with three conditions that manipulate the robot image, ranging from low to high human-like appearance, and we prompted participants to describe the robots and to give explanations they would expect to receive from the robot. 
Both descriptions and explanations were categorised and analysed.
The \emph{primary research question} for this paper is: Do robots with different levels of human-like appearance elicit different expected explanations and, if so, is there a difference in eliciting anthropomorphic vs non-anthropomorphic explanations? To contextualise our findings, we also set a \emph{secondary research question}: Are participant's descriptions of robots associated with humanlike appearance?

The contribution of this paper is twofold. First,  this empirical investigation is relevant to explainable robotics by providing evidence of the explanations users would expect from robots of different appearances. This will enable future work on explainable robotics to consider the mental model users may have depending on the robot human-like appearance when robots are to explain themselves. Second, our work contributes to understanding how the human-like appearance of robots influences how they are perceived, specifically in terms of mind attribution.

Our findings indicate that most explanations across conditions reveal the robot's mental capacities, showing that all the stimuli elicited a high level of anthropomorphism.
However, the level of human-like appearance is positively correlated with anthropomorphic explanations (referencing mental capacities) and reversely, there is a negative correlation between human-like appearance and non-anthropomorphic explanations. The attributed mental capacities are perception, desire and thinking, while the capacity for feelings or for making a choice was rarely cited.

\section{Background and Related Work}
Robots' appearance is known to influence how users perceive and interact with robots. In particular, a robot's appearance influences the perceived characteristics and capacities, such as autonomous behaviour \cite{Haring2013} intelligence \cite{haring2016influence}, perceived trustworthiness and empathy \cite{Zlotowski2016}, perceived expressiveness, usefulness, durability and smoothness \cite{crowell2019} and likeability \cite{trovato2016hugging,haring2016influence}. In addition to perceptions of the robot, the robot's appearance is also related to users' attitude towards interacting with the robot, such as willingness to touch the robot \cite{Haring2013}, feelings of hugging a robot \cite{trovato2016hugging} and willingness to donate money \cite{ranhee2014}.
Notably, the robot appearance \cite{haring2016influence} and particularly the level of human-like appearance \cite{Martini2016,Abubshait2021,looser2010} is associated with mental state attribution. Evidence suggests that with increasing human-like appearance, people tend to attribute more mental capacities to the robots \cite{looser2010}, however, this relationship might not be linear \cite{Martini2016}. Moreover, many studies investigating the relationship between human-like appearance and attribution of mental faculties often rely on participant's explicit judgement of mindedness which captures the conscious belief that the robot is endowed with mind 
\cite{Banks2021}, and it is less common to deploy implicit measures \cite{Banks2020,Banks2021}. What is more, and to the best of our knowledge, the specific relationship between the human-like appearance of robots and the implicit perception of mental capacities has not been studied in the context of expected robotic explanations.

To investigate whether and to what extent mental capacities attribution are revealed in the expected explanation as a function of human-like appearance, we use the multidimensional conceptual framework of anthropomorphism proposed by K\"{u}hne and Peter \cite{Kuhne2023}. 
There is a conceptual ambiguity around anthropomorphism with several existing definitions \cite{Kuhne2023}, but \cite{Kuhne2023} define anthropomorphism as the attribution of human mental capacities to robots, which they argue is an element commonly included in earlier conceptualisations of anthropomorphism, and thus reflects an agreement in the community regarding the centrality of this concept for anthropomorphism.
This conceptualisation clearly demarcates anthropomorphism from an \emph{upstream cognition}, which are the precursors of anthropomorphism, and \emph{downstream cognition}, which are the potential consequences of anthropomorphism \cite{Kuhne2023}. A notable distinction between some previous conceptualisations of anthropomorphism \cite{zhang} and the one at hand is the role of the robot's human-like appearance and movement. While these aspects have often been regarded as an aspect of anthropomorphism, the authors classify these factors as the precursors of anthropomorphism, meaning that human-like appearance and movement can elicit the perception that the robot has human mental capacities \cite{Kuhne2023}, which is supported by empirical research \cite{looser2010,Martini2016}. A consequence of anthropomorphism of robots is the attribution of personality and moral judgment, empirically supported by \cite{chee2012personality, Malle2016moraljusgement}, even though these studies considered appearance, rather than mind perception. The conceptualisation of anthropomorphism as proposed by \cite{Kuhne2023} comprises five dimensions based on the Wellman's framework of the Theory of Mind (ToM) \cite{wellman1990child} in \cite{Kuhne2023}. These dimensions -- thinking, feeling, perceiving, desiring and choosing (between the alternative courses of action) -- capture the five mental capacities in humans derived from ToM \cite{Kuhne2023}. In this study, we partially based our codebook on the framework by K{\"u}hne and Peter to identify and categorise anthropomorphic explanations provided by participants in the experiment described next.

\section{Method}

This between-subject study had three conditions, each presenting an image of a domestic service robot with a different level of human-like appearance. In an online survey, participants were asked to describe the robot and provide explanations what they would expect to receive from the autonomous domestic service robot whose role is to help with domestic chores.
Both descriptions and explanations were coded and examined for correlations with the human-like appearance of the robots.

\subsection{Visual Stimuli -- Robots}
\begin{figure}[t]
    \includegraphics[width=.4\textwidth]{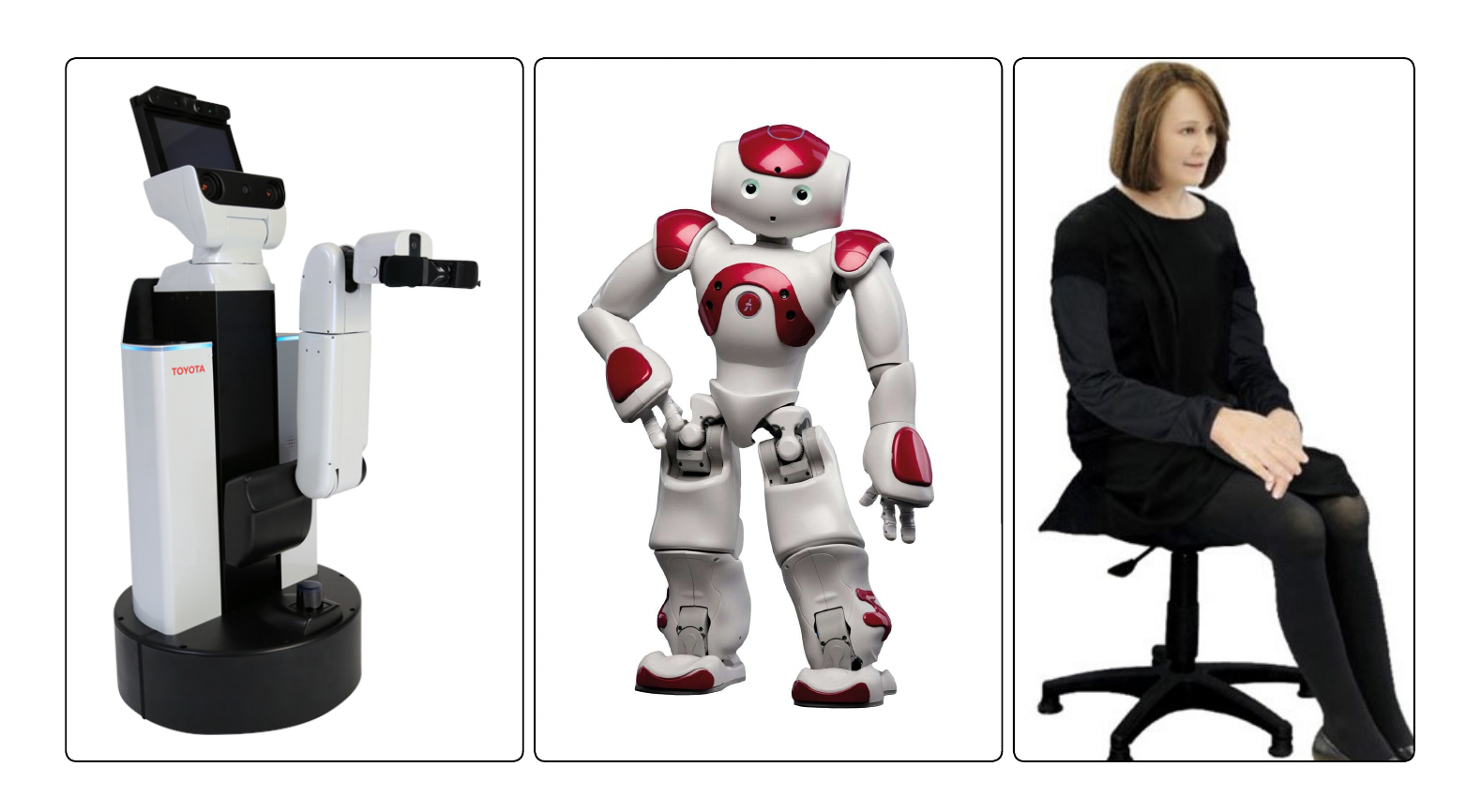}
    \caption{Our three conditions of robot appearance from ABOT ~\protect\cite{Phillips2018}.}
    \label{fig:robots}

\end{figure}

This study consists of three conditions  differentiated by the level of human-like appearance of the robot. For each condition, we selected an image of one robot with low, medium, and high level of human-like appearance, as this allowed us to consider the ordered nature of the variable while keeping the number of participants needed in the experiment manageable.
The robots images, shown in Fig.~\ref{fig:robots}, were selected from the well-known Anthropomorphic
Robot Database (ABOT\footnote{https://www.abotdatabase.info}) \cite{Phillips2018}.

ABOT is a collection of 251 real-world images with an overall human-likeness score and a score for the four dimensions of human-likeness: body-manipulators, surface look, facial features and mechanical locomotion \cite{Phillips2018}.  
For the condition of a robot with a low human-like appearance (further referred to as \lo), the Human Support Robot (HSR) by Toyota was chosen. According to ABOT, the human-likeness score of HSR is only 9.85 out of 100 \cite{Phillips2018}, so very low in human-like appearance. The HSR is equipped with an arm with a gripper, which could enable the robot to perform the domestic chore tasks introduced to the participants. 
Nao, developed by Aldebaran, was used as a robot of medium human-like appearance (further referred to as \mi), scoring 45.92/100 on the human-like appearance \cite{Phillips2018}, while also seemingly possessing all actuators to enable it to complete all the presented tasks.
Finally, Nadine, developed by Nadia Magnenat Thalmann (University of Geneva) was used for the condition of a highly human-like-looking robot (further referred to as \hi). Nadine has a human-likeness score of 96.95/100, currently the most human-like in  ABOT \cite{Phillips2018}.

\subsection{The domestic service robot and its tasks}
In all conditions, the robots are introduced as autonomous domestic service robots whose task is to assist the user with household chores. Users are then introduced to actions performed by the robot and are prompted to give an explanation they would expect to receive from the particular robot following the action. 
We adapted the set of actions from \cite{kopecka2024ijcai}, but we eliminated those items that might be perceived to be impossible to complete by the depicted robots. For example, the item `The robot made the bed' was excluded from this study, because some participants might find the \lo~robot to lack the actuators to complete the task, which might influence the results. 
The resulting set of actions is a comprehensive and diverse collection of domestic chores that would be familiar to participants.
The list of all actions used in our study can be found in the supplementary materials\footnote{\href{https://osf.io/8s2ej/overview?view_only=51898b8f70ae4a4c85e484b243b8976a}{Supplementary materials}}.

\subsection{Procedure}
This study was registered at the IRB at our institution as a study with minimal ethical risk. The questionnaire was scripted in Qualtrics\footnote{https://www.qualtrics.com} and administered on Prolific\footnote{https://www.prolific.co/} in November 2023. Before the full data collection, we ran a pilot study (N=5) for the \lo~condition, after which we changed the wording of the prompt to elicit the expected explanation from `How should the robot explain this action' to `Why did the robot do it?,' since the original formulation elicited some utterances that were not explanations (i.e. for the action “The robot suggested a popular cafe nearby.” participant submitted “Try this cafe”, which is not an explanation for the action). After this change in the wording, an additional pilot was run for all conditions, each with five participants, after which no further changes were made. 
Note that the instruction `Please provide an explanation you'd expect to receive from the robot explaining why it performed the stated action' was provided before each action, which remained the same in the pilot and the final version of the questionnaire. All actions are listed in the supplementary material.
Three versions of the survey were run in parallel, each for one condition of the study (that is one version per \lo, \mi, and \hi~human-like looking robot). The conditions were different only by manipulating the image of the robot displayed throughout the survey, while the verbiage of all the texts within the survey remained the same.
During the survey, participants were first provided with a participant information sheet and asked to express their consent to participation. Consenting participants were shown a large image of one of the robots, depending on the condition, and a description of the robot and its duties as a domestic service robot. To ensure participants' engagement with the image of the robot, we asked them to write three words that best describe the robot. Following this introductory task, participants proceeded to the main task of the study, which was the elicitation of expected robot explanations. Each participant was presented with randomly selected 10 out of 26 actions to be explained. Each action was displayed individually on a new page, along with the image of the robot and the instructions. Participants could leave feedback at the end of the survey. 

\subsection{Data Quality \& Participants}
We recruited 180 participants in total, 60 per condition. The inclusion criteria for this study were (a) residence in the UK, to limit a possible influence of culture and language 
and (b) being a reputable (approval rate \textgreater95\%) and experienced (\textgreater100 of previous submissions) participant on Prolific. The latter are known measures for increasing data quality \cite{peer2014reputation}. All questions in the questionnaire were open free-text questions and manual data quality checks did not identify any low-quality answers, such as nonsensical answers or extremely short (e.g. one-word) answers where explanations were requested. The mean age of participants is 42.75 years ($SD=14.70$, $min=20$, $max=78$), the sex composition of the sample is 72\% of females and 28\% of males and an Undergraduate degree is the most frequent educational attainment by our participants (37\%).

\section{Codebook development and analysis}
\label{sec:code_def}
\begin{figure*}
    \includegraphics[width=\textwidth]{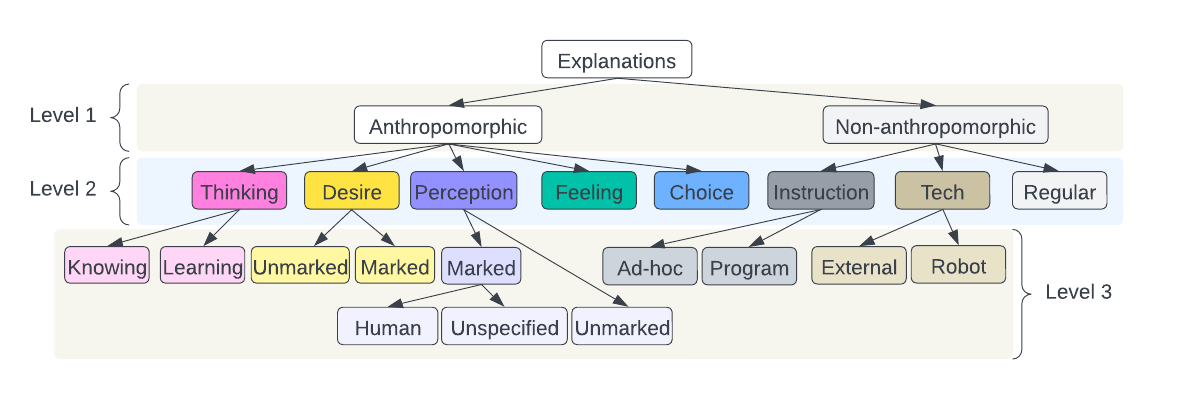}
    \caption{Hierarchical coding frame used to code explanations, partially based on~\protect\cite{Kuhne2023,Malle1999,degraaf2019} and partially inductively derived from the data. }
    \label{fig:codes}

\end{figure*}

Across the three conditions with the 180 participants, we collected 1800 explanations of robot actions (600 explanations per condition). These explanations were categorised using a mixture of deductive and inductive coding, organised in a hierarchical coding frame \cite{richards1995using}, represented by Fig.~\ref{fig:codes}. The initial codebook comprised two central codes in the first level of the hierarchy (L1): (1) anthropomorphic and (2) non-anthropomorphic explanations. The anthropomorphic explanations on Level 2 were informed by the dimensions of anthropomorphism proposed by K{\"u}hne and Peter \cite{Kuhne2023} and further expanded on level 3 (L3). The L3 categories for desire are informed by \cite{Malle1999} and the rest of the anthropomorphic L3 categories are inductively derived from the data. The L1 non-anthropomorphic code was inductively expanded by adding three codes into level 2 (L2): instruction, regular and tech (as detailed later) and the codes instruction and tech are further expanded into L3. L3  category program is informed by \cite{degraaf2019} and the rest of the L3 categories are also derived inductively from the data. We now describe the resulting anthropomorphic and non-anthropomorphic codes used. The whole codebook can be found in the supplementary materials to aid the transparency and reproducibility of our work.

\subsection{Anthropomorphic explanations}
    \textbf{(1) Thinking.} Explanations that imply the robot's capacity to think are those referring to any process of higher cognition such as reasoning, learning and remembering. It also includes the product of these processes, such as having beliefs \cite{Kuhne2023}. In our data, these often represent the robot's knowledge and ability to learn:

        \underline{(1a) Knowing.} The robot has access to some knowledge or beliefs that guides its behaviour. . 
        For example: \textit{Because the robot knows the users temperature preferences (Condition: \lo, question: Q10 (in supplementary materials)). }

        \underline{(1b) Learning.} This  comprises explanations  in which the robot learnt its behaviour, typically by observing human behaviour and either learning to imitate it or learning the human’s shortcomings and compensating for them. For example: \textit{The robot has learnt that at a certain time each day, a coffee is made (\hi, Q2).}

    \textbf{(2) Desire.} Explanations in this category indicate the capacity of the robot to have desires, needs and preferences \cite{Kuhne2023}, referring to the desires to be fulfilled by the action the robot is explaining. Following \cite{Malle1999,Fex},  
    we further categorise these explanations according to the presence of linguistic markers:
    
        \underline{(2a) Marked.} The linguistic markers for desires are typically \textit{want} and \textit{need} \cite{Fex}. An example of a marked desire reason is: \textit{The robot \underline{wants} to look after the user's health (\lo, Q23).}
        
        \underline{(2b) Unmarked.} Unmarked desires directly state the desire without the linguistic marker. For example, \textit{To keep the plant healthy (\hi, Q16).}

    \textbf{(3) Perception}. The robot's explanation references its ability to perceive its environment and thus references external stimuli to the robot \cite{Kuhne2023}. Here we also distinguish further categories depending on whether the explanation contains an explicit reference to the particular sensory system that facilitated the perception:

        \underline{(3a) Unmarked.} Unmarked perception explanations contain information about the robot's environment and other agents that the robot can perceive, but it does not state how the robot perceives it. An example of such an explanation is \textit{The fridge was empty (\hi, Q6).}
        
        \underline{(3b) Marked-human.} These explanations 
        mention how the robot accessed the environment and it explicitly references a human sense (vision, hearing, touch, taste and smell). For example, \textit{It saw creases in the fabric (\hi, Q24).} 

        \underline{(3c) Marked-unspecific.} These explanations also reference how the robot accessed the environment, but they do not cite human senses.
        The markers considered as unspecific include: noticed, perceived, observed. For example, \textit{The robot noticed the fuel tank was low (\mi, Q17).}

    \textbf{(4) Feeling.} These explanations reference the robot's capacity to have feelings, 
    either bodily experiences (i.e. pain), or emotions (i.e. fear) \cite{Kuhne2023}, e.g., 
    \textit{The robot was fed up of being inside (\hi, Q25)}.
    
    \textbf{(5) Choice.} These explanations reference the robot's ability to choose between alternatives \cite{Kuhne2023}, e.g., \textit{I hung the wet washing to dry instead of using the tumble dryer,  to save money (\lo, Q9).}

\begin{figure*}[t]
    \centering
    \includegraphics[width=\textwidth]{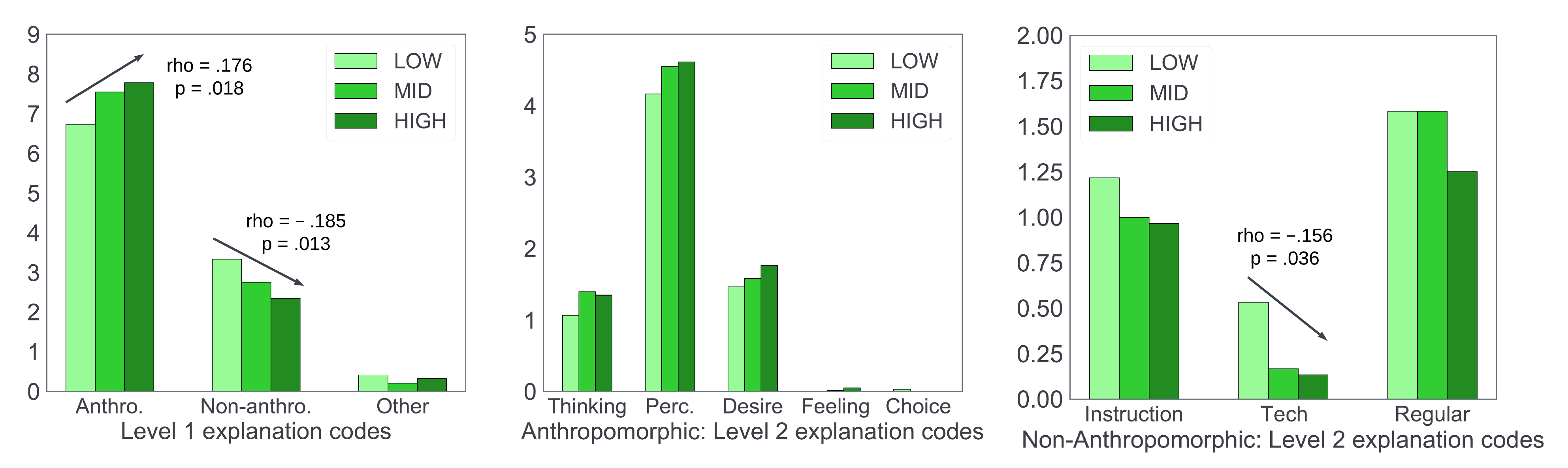}
    \caption{Results of L1 and L2 codes correlations. Only statistically significant results are annotated.}
    \label{fig:results_l12}

\end{figure*}

\subsection{Non-anthropomorphic explanations}

    \textbf{(6) Instruction.} These explanations reference that the robot was instructed (by the programmer or the user) to perform the given action, which are further categorized into:
    
        \underline{(6a) Ad-hoc.} These are instructions or requests typically given to the robot by the user on the fly. As such, the capacity to follow instructions supplied by the user implies potential for user interaction. An example of such an explanation is \textit{Because the user asked for recommendations (\lo, Q13).}
        
        \underline{(6b) Program.} These are instructions the robot is programmed with. This category contains all explanations in which the robot refers to its programming or software, as previously proposed by \cite{degraaf2019}. For example, \textit{Because it’s been programmed to separate lights and darks (\hi, Q1).}
    
    \textbf{(7) Regular.} These explanations  reference some recurring event, or condition that comes up with some regularity.
    An example is \textit{It was a regular meal time (\lo, Q7).}
    
    \textbf{(8) Tech.} These explanations  reference some technological way of receiving information, and can be further categorized depending on the source:
    
    \underline{(8a) External.} The robot receives information from other (smart) appliances, which is used to explain the action. 
    For example, \textit{It was linked to the user's smartwatch and the user had been stationary for an amount of time (\mi, Q21).}
    
    \underline{(8b) Robot.} 
    Conceptually, these explanations are similar to perception, but they use machine-specific language (e.g. scanned, used a sensor), which can be interpreted as evidence against anthropomorphism, e.g.
    \textit{It was dark and it's sensors were triggered (\lo, Q22).}

\subsection{Data Coding and Analysis}

After coding the 1800 free text entries, 600 per condition, our dataset contained 1888 coded explanations, as some entries contained more than one element and were thus assigned to more than one code. All explanations were coded by the author, and half of the explanations (N=900) were coded independently by another researcher to check for reliability. The inter-rater reliability, measured using Cohen's $\kappa$, suggested high agreement between codes assigned by the two researchers independently for L2 ($\kappa=.74$) as well as for L3 ($\kappa=.72$).  
The number of coded explanation elements is similar across conditions (\lo=629, \mi=631, \hi=628).

In order to analyse the association between the human-like appearance of robots and the expected explanations, we calculated a set of Spearman's rank-order correlation, one for each code, between the three conditions (\lo, \mi~and \hi~human-likeness) and a count of given explanation per respondent. We chose a correlational study to account for the ordered nature of our conditions, thus we test for a monotonic relationship between the variables. As this is an exploratory study, the results are not corrected for multiple testing \cite{BENDER2001} and we consider the standard $p=.05$ as a threshold for statistical significance.

\section{Results}
\subsection{Explanations and human-like appearance}

In the first level of analysis (L1), we investigated whether the level of human-like appearance is correlated with anthropomorphic and non-anthropomorphic explanations in general.
Our results show that with increasing levels of human-like appearance of the robot, people expect to receive more anthropomorphic explanations ($\rho=.176, p=.018$) and, inversely, fewer non-anthropomorphic explanations ($\rho=-.185, p=.013$). However, it should be noted that in all three conditions, between \(\frac{2}{3}\) and \(\frac{3}{4}\) of explanations were anthropomorphic (\lo= 64\%, \mi= 72\%, \hi=74\%). Explanations that were not categorised either as anthropomorphic or non-anthropomorphic, constituting only 3\% of all explanations, were not associated with the robot's appearance ($\rho=-.034, p=.647$).

Although the robot's appearance shows a monotonic relationship with the aggregated category (L1) of anthropomorphic explanation, neither of the more specific categories of anthropomorphic explanations (L2), thinking ($\rho=.134, p=.071$), perception ($\rho=.075, p=.316$), or desire ($\rho=.061, p=.416.$),  were correlated with the robot's appearance. Feeling and Choice explanations were rare in our data and
while they were included in anthropomorphic explanations, they were not individually assessed for correlation.
Only four and two explanations were classified as a feeling and choice explanation, respectively. Regarding L2 codes for non-anthropomorphic explanations, \textit{tech} is negatively correlated with human-like appearance ($\rho=-.156 , p=.036$), but no association was identified for the codes instruction ($\rho=-.126, p=.091$) and regular ($\rho=.065 , p=.381$).
The average count of explanation type per participant in Levels 1 and 2 provided per condition is represented in Fig.~\ref{fig:results_l12}.

An inspection of Level 3 of the hierarchical coding frame did not reveal any correlations between any of the anthropomorphic categories.  The categories of thinking, which are knowing ($\rho=.113, p=.130$) and learning ($\rho=.088, p=.240$), desire with the categories unmarked ($\rho=.039, p=.601$) and marked ($\rho=.021, p=.778$), and perception, with the categories unmarked ($\rho=.084, p=.260$), marked-human ($\rho=.003, p=965$) and marked-unspecific ($\rho=.038, p=.610$) are not correlated with human-like appearance.

\begin{figure*}[t]
    \centering \includegraphics[width=1\textwidth]{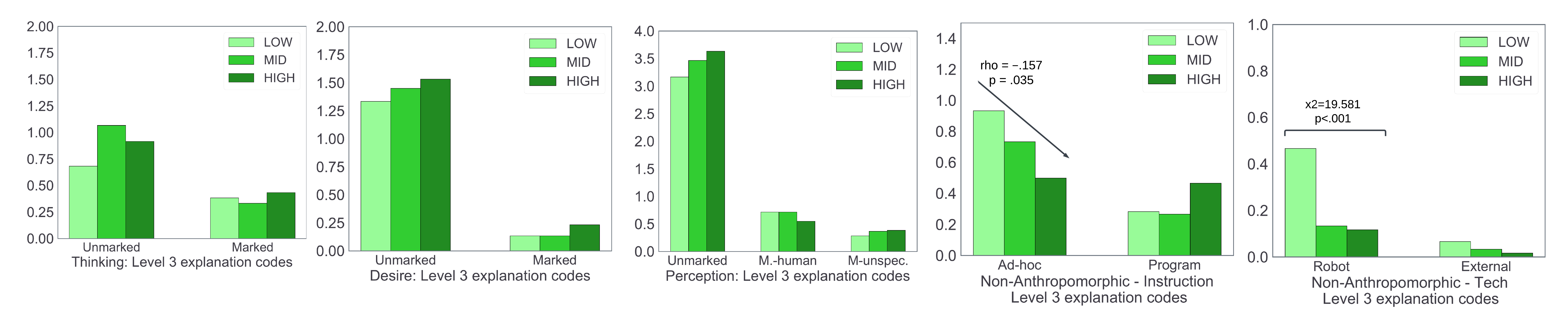}
    \caption{Results of L3 codes correlations. Only statistically significant results are annotated.}
    \label{fig:results_l3}

\end{figure*}

Next, we report on the non-anthropomorphic codes on Level 3. 
An investigation of the two kinds of codes constituting the code instruction on level 3 reveals an interesting finding: while the expectation to explain their behaviour by citing program is the same for the three conditions \lo, \mi~ and \hi~($\rho=.0370, p=.621$), the expectation that the robot would refer to an ad-hoc instruction by the user decreases with higher human-like appearance of the robot ($\rho=-.157, p=.035$). 
Although we did not identify a monotonic relationship between any categories of tech, robot ($\rho=-.128, p=.086$) and external ($\rho=-0.106, p=.158$), it is worth noting that additional one-way Chi-Square test (${\chi}^2(2)=19.581, p<.001$) indicated that robot's human-like appearance and explanations in the code tech-robot have significantly different frequencies between at least two conditions. By visually analysing the frequencies per condition and confirming by performing three pairwise one-way Chi-Square test, the \lo~condition has significantly higher frequencies of the tech-robot explanations than both \mi~and \hi~conditions. The frequencies of explanation codes on Level 3 per condition are represented in Fig.~\ref{fig:results_l3}.

\subsection{Robot descriptions and human-like appearance}\label{sec:descriptions}
To examine trends in how people describe robots, we categorised all the descriptions into the following categories: (1) human characteristics, e.g. friendly, obedient, understanding; (2) describing the robot explicitly as humanlike or lifelike; (3) robotic characteristics, e.g. hitech, programmed, digital; (4) explicitly describing the robot as robotic; (5) mentioning gender, e.g. female, woman, lady; (6) eerie, which also includes descriptors such as unsettling, uncanny and similar; (7) social role, which are exclusively descriptors of the robot as an aid, e.g. helper, housekeeper, servant and nanny; (8) physical characteristics, e.g. tall, shiny and ugly; (9) intimidating, including dangerous and similar; (10) autonomous or independent; (11) intelligent, smart or clever; (12) useful or helpful. Our data indicate a positive correlation between the robot's human-like appearance and being described with human-like descriptors or explicitly described as humanlike, eerie, and being ascribed gender. On the other hand, a robot's human-likeness is negatively correlated with robot-specific descriptions (both explicitly and robotic characteristics) and physical description. Being described as intimidating, autonomous, intelligent, useful or by social role is not correlated with the human-like appearance of the robot. For correlation coefficients see Tab.~\ref{tab:descriptions}.

\begin{table}[!t]
    \centering
    \begin{tabular}{rl|ccc|l}
    
         & & \lo~(N) & \mi~(N) & \hi~(N) &           $\rho$ \\ 
         \toprule
        1&Human char. & 17 & 38 & 44 &    ~.156* \\ 
        2&`Human-like' & 0 & 3 & 13&       ~.296***\\ 
        3&Robotic char. & 39 & 24 & 17  &  --.233**\\ 
        4&`Robotic' & 4 & 1 & 0 &         --.166*\\ 
        5&Gender (fem.) & 0 & 0 & 15 &               ~.369*** \\ 
        6&Eerie & 3 & 3 & 14  &            ~.226**\\ 
        7&Physical char. & 37 & 32 & 8  &  --.275***\\ 
        8&Social role & 2 & 12 & 10 &                
        ~.12\\ 
        9&Intimidating & 3 & 0 & 1 &                 --.092 \\ 
        10&Autonomous & 12 & 12 & 10  &                         --.020 \\ 
        11&Intelligent & 12 & 5 & 9  &                          --.025 \\ 
        12&Useful/helpful & 29 & 19 & 20  &                     --.141 \\ 
        13&Other & 22 & 31 & 19 &  \\ 
    \end{tabular}

    \caption{Correlation between descriptions and human-like appearance. *$<.05$, **$<.01$, ***$<.001$. Char. = Characteristic.}

    \label{tab:descriptions}

\end{table}

\section{Discussion}
Overall, our results suggest that a human-like appearance elicits a higher level of anthropomorphism in expected robot explanations, and in the robot descriptions. Next, we discuss our main take-aways. 

\subsection{High rates of anthropomorphic explanations}
Our results identified differences in the expected robot explanations according to the robot's human-like appearance. However, in all three conditions, between 2/3 and 3/4 of explanations comprised anthropomorphic explanations. 
This partially differs from the findings of \cite{Martini2016}, who studied the relationship between human-like appearance and explicit mind attribution, without considering the robot's capability or behaviour, and found that people attribute mental capacities to robots only after the robot's appearance reaches a certain degree of human-likeness, after which an increase in human-likeness also increases mind attribution. The fact that our findings indicate a high level of mind perception for all conditions, even the \lo~one, could be explained in at least two ways. 
First, in addition to \textit{ appearance}, the robot's \textit{behaviour} is known to influence anthropomorphism \cite{Kuhne2023,cucciniello2023mind,wallkotter}, so the high, baseline level of anthropomorphic explanations across all conditions could be 
due to introducing the robot as autonomously operating in the household and providing explanations,
which would require some capacity of information processing (\(\approx thinking\)), perception, planning and goal-orientedness (\(\approx desiring\)).

Second, we did not explicitly ask participants whether they believe that the robot has mental capacities (as done in \cite{Martini2016}), but we studied whether participants reveal their perception of the robot's mental capacities in the explanations. The difference between our findings and \cite{Martini2016} 
could be understood as the difference between explicit and implicit attribution of mental capacities \cite{Koban2024}.
Since implicit attribution of mental capacities is automatic and subconscious, and explicit attribution of mental capacity is controlled and deliberative \cite{Banks2020,low2012implicit}, the results of \cite{Martini2016} and our study considered together could mean that even the \lo~robot is subconsciously and spontaneously perceived to have mental capacities, but when explicitly asked, people are apprehensive to attribute mental capacities to robots until a certain point of human-like appearance.

\subsection{Anthropomorphic, but no Feelings or Choice}
 
The majority of anthropomorphic explanations across conditions are perception, desire and thinking. Notably, very few explanations cited either the robot's feeling or choice. 
There is evidence that people find it eerie when machines express emotions \cite{Stein2017,Gray2012,Appel2020} and while people tend to attribute a degree of agency (the capacity to act and think) to robots, much less so the capacity to have experience (to feel and sense) \cite{Gray2007}. 
Another reason for the prominence of thinking, perception and desire in robot explanations is that these concepts can be mapped onto  beliefs, desires and causal history of reasons (CHR) which are mental states used to explain intentional actions of people as well as robots \cite{degraaf2019}. 
The formal analysis of correspondence between these two sets of categories (thinking, perception and desire in our study and beliefs, desires and CHRs, as defined in~\cite{degraaf2019})
could be an avenue for future research. 
Regarding choice explanations, 
Miller \cite{Miller2019} posits that explanations are \textit{contrastive}, i.e., rather than addressing the question of \textit{why the robot did X}, they would address \textit{why the robot did X instead of Y}, even if the alternative action is not explicitly mentioned. In our data, only a negligible number (N=2) of explanations explicitly mentioned the alternative course of action. 
However, such explanations
were found to yield a better understanding of the situation, e.g., in  autonomous vehicles \cite{Omeiza2021}, and should be considered by future research. 

\subsection{Senses vs. sensors: when `perception' does not imply anthropomorphism}
In non-anthropomorphic explanations, 
Tech was the only code that correlated, negatively, with human-like appearance. Further inspection of L3 codes revealed that tech-robot is over-represented for the \lo~robot. 
To remind the reader, tech-robot are explanations referencing the robot's ability to access its environment in a machinic way (e.g. scanning or using a sensor). 
Although tech-robot explanations are conceptually close to perception explanations, we classify them as non-anthropomorphic, given that not all instances of \textit{sensing and reacting} to the environment are indeed instances of perception \cite{boden1969machine}. In fact,  \cite{boden1969machine} argues that perception requires a sufficient level of behavioural complexity, autonomy and voluntariness. 
For example, a sea anemone is sensitive to light and reacts to light by changing its muscular tonus, but such behaviour does not necessarily imply perception \cite{boden1969machine}.
For the explanations in the tech-robot category, participants specifically chose to use machinic channels of accessing the environment (using sensors, detecting or scanning), which is more akin to \textit{measuring} the environment than to \textit{perceiving} it and  
we suspect that this might be a strategy to \textit{resist} anthropomorphising the robot by focusing on the machinic aspect despite understanding that the robot can access and react to its environment. This is particularly prominent in the \lo~condition according to our data, suggesting the influence of the robot's appearance on opting for explanations based on the robot's machining manner to access its environment.

\subsection{User-supplied instructions are inversely correlated with human-likeness}

In our results, with an increasing level of human-likeness, participants cited fewer ad-hoc instructions from the user in their explanations. We propose two possible reasons. 
First, this might be caused by the uncanny valley, a phenomenon describing the effect of uneasiness elicited by robots with high human-likeness \cite{mori1970uncanny}. 
We observe indicators of the uncanny effect in the robot descriptions we collected, where eeriness is positively correlated with human-like appearance.
This might lead to robot anxiety, which is associated with avoiding communication with the robot \cite{Nomura2008}.  
As a result, people might feel unwilling to communicate their instructions to the robot, so 
they would not expect the robot to explain themselves citing those instructions. 
The second reason could be due to people perceiving the robots as having a different social role. In this way, lower human-likeness would elicit a perception of the robot as a utilitarian tool, which can be told to perform chores, but with increasing human-likeness, the robots are perceived as social entities whom people are more reluctant to treat as tools to perform work for them. From the robot descriptions, we know that with increasing human-likeness, people tend to describe the robots more often using human characteristics and less with robotic characteristics.
These two hypotheses are not mutually exclusive as robots introduced as tools elicit less eeriness than robots introduced as having agency and a capacity for experiencing \cite{Appel2020}.

\subsection{Limitations and future research}
In this final section of this paper, we acknowledge some limitations of this work and propose several directions for the future research.
The independent variable investigated in this study is the overall human-like appearance, represented by three conditions each with a robot of different degrees of human-like appearance. However, some other aspects of the robots' appearance not investigated in this study could influence the results. For example, \cite{zhao2019people} found that specific aspects of a robot's human-like appearance, such as body-manipulators or face, contribute to the attribution of different mental capacities and this study could be extended in the future to investigate these in detail. 
Furthermore, humanlikeness is not affected only by appearance and it would be interesting to investigate other factors increasing human-likeness, such as conversational capacities, akin to those of ChatGPT \cite{jacobs2023brief, zhan2023deceptive} or the use of synthesised speech/voice \cite{weitz2021let} instead of text.

 The next direction for future research that we propose here is the further examination of the relationship between the robot's human-likeness and explanations to understand different aspects of this relationship. We studied the effect of a robot's appearance on the explanations people would \textit{expect} to receive, however, it is not clear whether that aligns with an explanation people would \textit{prefer} to receive or find the most \emph{useful}.

 A known problem in social robotics is the `perceptual belief problem' described by Thellman and Ziemke~\cite{Thellman2021}, as the mismatch between the actual perceptual and cognitive capacities of the robot and what people think are the perceptual and cognitive capacities of the robot~\cite{Thellman2021}. It would be interesting to investigate whether different kinds of explanations could also be used strategically to close this gap between the user's perception and the actual robot's capacities either by providing its perceptual beliefs to help the user to calibrate their mental model of the robot, as suggested by~\cite{Thellman2021}, but also by using less anthropomorphic explanations in situations where users might perceive the robot as possessing more advanced perceptual and cognitive skills then it actually does. 
 Given the known role of the robot's appearance in the perception of the robot's abilities, which is also evident from our work, this might be particularly useful for robots with a high level of human-like appearance if it elicits an inadequately high perception of the mind.

Next, the participant sample recruited in this study was drawn from the UK and data collection was conducted in the English language, because it is known that several aspects of human-robot interaction \cite{lim2021social}, robot perception \cite{Castelo2022} and acceptance \cite{Brohl2019} vary between nations. Therefore, we advice not to extrapolate our results to other national contexts. Replicating our study in other national contexts and comparing the results with ours is another exciting avenue for future research.
Finally, a confirmatory study should be conducted to confirm the findings of this exploratory study.



\section*{ACKNOWLEDGMENT}
We thank Elfia Bezou-Vrakatseli for coding portion of the explanations, Lennart Wachowiak for his feedback on the research instrument and all the participants.

\bibliographystyle{IEEEtran}

\bibliography{references}

\end{document}